\documentclass[review]{elsarticle}

\usepackage{amssymb}
\usepackage{amsmath}

\usepackage{url}
\usepackage{booktabs}						
\usepackage{multirow}						
\usepackage{tabularx}
\usepackage{amsfonts}						
\usepackage{graphicx}						
\usepackage{duckuments}						
\usepackage{algorithm}
\usepackage{algpseudocode}
\usepackage{tikz}
\usepackage{array}
\usepackage{pifont} 
\usepackage{pgfplots}
\pgfplotsset{compat=1.18}
\usetikzlibrary{arrows.meta, positioning, shapes.geometric}
\usepackage{xspace}
\usepackage{caption} 
\newcommand{\todo}[1]{\textcolor{blue}{#1}}
\newcommand{\REM}[1]{\textcolor{blue}{}}
\newcommand{\name}[0]{HEHRGNN\xspace}
\newcommand{\kgname}[0]{KG\xspace}
\newcommand{\kgnames}[0]{KGs\xspace}
\newcommand{\cmark}{\ding{51}}%
\newcommand{\xmark}{\ding{55}}%

\journal{Knowledge-Based Systems}

\begin{document}

\begin{frontmatter}


\title{HEHRGNN: A  Unified Embedding Model for Knowledge Graphs with Hyperedges and Hyper-Relational Edges}


\author{Rajesh Rajagopalamenon, Unnikrishnan Cheramangalath} 

\affiliation{organization={Department of Computer Science and Engineering\\ Indian Institute of Technology Palakkad},
          country={India}}

\begin{abstract}
        Knowledge Graph(KG) has gained traction as a machine-readable organization of real-world knowledge for analytics using artificial intelligence systems.
Graph Neural Network(GNN), is proven to be an effective \kgname  embedding technique that enables various downstream tasks like  link prediction,  node classification, and graph classification. The focus of research in both \kgname embedding and GNNs has been mostly oriented towards simple graphs with binary relations.
However, real-world knowledge bases  have a significant share of complex and n-ary facts that cannot be represented by binary edges. More specifically, real-world knowledge bases are often a mix of two types of n-ary facts - (i) that require  hyperedges and (ii)  that require hyper-relational edges. Though there are research efforts catering to these n-ary fact types, they are pursued independently for each type.
We propose \underline{H}yper\underline{E}dge \underline{H}yper-\underline{R}elational edge \underline{GNN}(HEHRGNN), a unified embedding model for n-ary relational \kgnames\ with both  hyperedges and hyper-relational edges. The two main components of the model are i)HEHR unified fact representation format, and ii)HEHRGNN encoder, a GNN-based encoder with a novel message propagation model capable of capturing complex graph structures comprising both hyperedges and hyper-relational edges.
The experimental results of HEHRGNN on link prediction tasks show its effectiveness as a unified embedding model, with inductive prediction capability, for link prediction across real-world datasets having different types of n-ary facts. The model also shows improved link prediction performance over baseline models for hyperedge and hyper-relational datasets. 
\end{abstract}

\REM{ \begin{abstract}
 Knowledge Graph(KG) has gained traction as a machine-readable organization of real-world knowledge for use by analytics\todo{what analytics?} or artificial intelligence systems. Graph Neural Network(GNN), \REM{which has emerged as} an effective deep learning architecture for graph-structured data,  has proven to be an effective \kgname embedding technique that enables various downstream tasks like knowledge graph completion, link prediction, and node classification. The focus of research in both knowledge graph embedding and GNNs has been mostly oriented towards simple graphs with binary relations. However, real-world knowledge bases generally have a significant share of complex and non-binary facts that cannot be represented by binary edges. More specifically, real-world knowledge bases are often a mix of two types of non-binary facts - \REM{those} that require relational hyperedges and \REM{those} that require hyper-relational edges. Though there are research efforts catering to these non-binary fact types, they are pursued independently for each type. \REM{This paper brings out the need for unified research and models natively catering to both types of non-binary facts.} We  propose HyperEdge Hyper-Relational edge GNN(\name), a GNN-based unified model for inductive prediction on non-binary relational knowledge graphs, supporting both hyperedges and hyper-relational edges.  \name\  has a novel GNN architecture with a 4-step propagation mechanism forming an encoder capable of capturing complex graph structures comprising both hyperedges and hyper-relational edges.   The experimental results show the effectiveness of \name\ as a unified link prediction model, with inductive prediction capability, that can work across real-world datasets having different types of non-binary facts. The model also shows improved link prediction performances over benchmark datasets. \todo{experimental results should be  highlighted here, with numbers}
 
\end{abstract}
} 


\begin{highlights}
\item Unified Embedding Model for Knowledge graphs with Hyperedges and Hyper-Relational Edges
\item GNN Message propagation for Hyperedges and Hyper-Relational Edges
\item Link prediction on N-ary relational Knowledge Graphs
\item Unified fact representation format for N-ary knowledge graph datasets
\item Knowledge HyperGraphs and Hyper-Relational Knowledge Graphs 
\end{highlights}


\end{frontmatter}


\vspace{-0.2in}
\section{Introduction}\label{sec:Intro}
Knowledge Graph (KG) is the representation of a collection of connected facts as a  graph object.
The entities in the facts become the nodes  and the relations between the entities become the edges in a KG.
\REM{Knowledge representation, knowledge graph construction and knowledge graph completion are three different research areas  for a large knowledge base.}
Of late, KGs have become critical components that support Large Language Models(LLMs) in retrieval augmented generation by providing context-specific and up-to-date knowledge\cite{Xu2024Retrieval-augmentedAnswering}.
Knowledge Graph Completion(KGC), i.e., inferring missing facts from the  facts in a KG, is an essential task in KG. The collection of facts corresponding to even a specific domain could be innumerable, that it becomes practically infeasible to build a complete \kgname capturing all the facts. Inference on \kgnames\ by operating on the original high-dimensional data is a computationally hard task.  Hence, Knowledge Graph Embedding(KGE) that embeds the nodes and edges in a low-dimensional continuous vector space is used  as an effective technique for KGC. KGE is one of the important research area related to KGs.

Graph Neural Network(GNN), with its message passing mechanism capable of capturing the neighborhood structure and combining them with the node and the edge attributes,  has emerged as a promising approach for KGE and  link prediction  \cite{Schlichtkrull2017ModelingNetworks}, \cite{VashishthCOMPOSITION-BASEDNETWORKS},\cite{Wang2021RelationalCompletion}, \cite{NathaniLearningGraphs}. GNNs provide expressive power as well as inductive capability for prediction on graphs.

\par The  prior research works  that combines  KGE and GNN focus  mostly on  graphs with only binary edges that  connects  only two nodes. 
These binary edges can represent facts in the form of triples ( \texttt{h},\texttt{r},\texttt{t}) where entities \texttt{h} and \texttt{t} are connected through a relation \texttt{r}.  
However, real-world  knowledge bases generally have a significant share of complex n-ary facts that cannot be represented using  binary edges.
 The effective representation of n-ary facts in KGs  require \textit{hyperedges}\cite{huang2024link} \cite{fatemi2021knowledge} and \textit{hyper-relational edges}\cite{Rosso2020BeyondPrediction}\cite{galkin2020message}. 
A \textit{hyperedge}  connects two or more nodes. In the case of KGs, the term \textit{relational hyperedge} is used to indicate that there could be multiple relation types and each \textit{relational hyperedge} is labeled with a relation type(See Figure~\ref{fig:sample_HE_KG} in Section~\ref{sec:background}). A \textit{hyper-relational edge} consists of a   primary relational edge and an associated set of qualifier key value pairs for the primary relation (See Figure~\ref{fig:sample_HR_KG} in Section~\ref{sec:background}). 

Prior works in GNNs for n-ary relations like \cite{feng2019hypergraph}, \cite{bai2021hypergraph} cater to hyperedges with single relation type. 
There are  works catering to either Knowledge Hyper Graphs (KGs with relational hyperedges)\REM{\cite{YadatiNeuralHypergraphs},\cite{fatemi2021knowledge}, \cite{Vazquez2023KnowledgeTaslakian} \cite{huang2024link}} or Hyper-relational \kgnames(KGs with hyper-relational edges)\REM{\cite{Rosso2020BeyondPrediction},\cite{galkin2020message}} but not both. However, real-world knowledge bases often have a mix of these two types of facts. The current practice is to convert unsupported complex facts 
to a simpler form supported by the available models, leading to information loss and reduced expressivity. \emph{This  demands a unified model that supports both hyperedges and hyper-relational edges.} 

\par In this work, we propose \name,  a unified embedding model for  \kgnames\ with both hyperedges and hyper-relational edges. In addition to this, we attempt to bring more clarity on n-ary fact types required to be handled by \kgnames, their classifications and definitions. We also discuss the need for KGE models natively supporting these complex fact types. We believe that KG research should look at unified modeling paradigms natively supporting all fact types - binary and different n-ary types - for effectively capturing the real-world knowledge bases.  In fact, the broader definitions of \kgnames\ like "any graph-based representation of some knowledge can be considered a KG"\cite{paulheim2016knowledge} encompass all types of facts and relations. 

The major contributions of this paper are the following:
\begin{itemize}

\item Presents \underline{H}yper\underline{E}dge \underline{H}yper-\underline{R}elational edge \underline{GNN}(\name), a unified embedding model for KGs with both hyperedges and hyper-relational edges, comprising the two core components: 
\begin{itemize}
    \item HEHR, a unified fact representation addressing two aspects - the definition of format for representing facts in KG dataset files and the design of data structures for storing facts in memory for embedding generation. This format enables capturing binary  and n-ary  facts in various datasets without any loss of information.
    \item \name, a GNN-based encoder, with inductive capability, for link prediction on \kgnames with both n-ary facts types - hyperedge and hyper-relational edge facts. \name has a novel GNN message propagation model that forms the basis of the encoder component.
\end{itemize}
    
  \item An end-to-end encoder-decoder model for link prediction is implemented based on the proposed \name to validate the performance with respect to the hyperedge and hyper-relational benchmark datasets.  It shows an improved performance over baseline models. The effectiveness of the unified model is also validated by experiments on a combined dataset.  
\end{itemize}
To the best of our knowledge, this is the first work to propose a unified KG embedding model  capable of handling both hyperedges and hyper-relational edges directly.                                          
\vspace{-0.2in}
\section{Background}\label{sec:background}
\textbf{Knowledge Graph(KG):} A KG organizes and represents a collection of real-world facts in a machine-readable way using a graph data structure\cite{paulheim2016knowledge} 
\cite{BonattiKnowledge18371}. The entities in the facts form the nodes and the relationships among the entities form the edges in the KG.   \kgnames\ helps to provide the context and depth for analytics or artificial intelligence systems. 
A KG is formally defined in the literature\cite{Rossi2021KnowledgeAnalysis}\cite{Ji2022AApplications}, \cite{EhrlingerTowardsGraphs} and \cite{Hogan2021KnowledgeGraphs} as: 
$\;\mathcal{KG = \{V,R,F\}}$ where \( \mathcal{V}\) is the set of entities $v_i$; \( \mathcal{R}\) is the set of relations $r_j$; \( \mathcal{F}\) is the set of facts $f_k$. This definition assumes that each fact is binary i.e $\mathcal{F} \subset (\mathcal{V} \times \mathcal{R} \times \mathcal{V})$. The facts are written as triples $<h, r, t>$ where h is the head entity, r is the relation, and t is the tail entity. However, triple based paradigm is insufficient to capture complex real-world facts that have multiple entities and relationships.\\  
\textbf{N-ary relations or Non-binary relations:} Relations involving more than two entities are referred to in the literature using either of the two terms - \textit{n-ary relations} or \textit{non-binary relations}. \REM{The existing triple-based KG paradigms handle a n-ary fact by decomposing it into a set of binary facts. This decomposition results in either information loss or creation of additional virtual entities.}We stick to the term n-ary relations in this paper. \REM{ It is seen that the term \textit{n-ary} is sometimes used in the literature to refer specifically to facts or relations represented by hyperedges. However, we use it in the more commonly used  of any relation involving more than two entities} The n-ary relations or facts are represented using mainly two formalizations\cite{Wei2025AGraphs}: i)Hyperedges (or more specifically named Relational Hyperedges to indicate the multiple relation types in KGs), and ii)  Hyper-Relational edges.\\
\textbf{Hyperedge} is an edge that connects more than two nodes in a graph. A KG with hyperedges is  called a Knowledge Hypergraph. Its formal definition in the literature\cite{Ren2023NeuralDatabases} is similar to the \kgname definition above. The difference is that $\mathcal{F} \subset (\mathcal{R} \times \mathcal{P(V)} )$. The number of entities that are part of a hyperedge is called its \textit{arity}. There is an alternative formalization, referred in the literature\cite{guan2019link} \cite{Wei2025AGraphs} as role-value pairs, that uses a set of role-value pairs in place of a hyperedge relation. It can be seen that role value pairs formalization is equivalent to hyperedge formalization in terms of the facts that can be represented, and hence it need not be considered separately. 
\\
\textbf{Hyper-relational edge} comprises a primary tuple and an associated set of qualifier pairs.  A KG with hyper-relational edges is called a Hyper-relational KG. Its formal definition in the literature \cite{galkin2020message},\cite{Alivanistos2021QueryGraphs} and \cite{Ren2023NeuralDatabases} is similar to the \kgname definition above. The difference is that $\mathcal{F} \subset (\mathcal{V} \times \mathcal{R} \times \mathcal{V} \times \mathcal{P(R} \times \mathcal{V))}$ where $\mathcal{P}$ denotes the power set. A hyper-relational fact is written as $(s, r, o, \mathcal{Q})$, where (s, r, o) is referred to as the primary tuple of the fact and $\mathcal{Q}$ is the set of qualifier pairs $\{(qr_i, qv_i)\}$ with qualifier relations $qr_i \in \mathcal{R}$ and qualifier entities $qv_i \in \mathcal{V}$.\\~\\
We refer to the facts which can be represented directly by a hyperedge as \textit{hyperedge facts} and facts which can be represented directly by a hyper-relational edge as \textit{hyper-relational facts}. Unlike a hyper-relational edge that has a primary relation and a set of qualifier relations, a hyperedge has a single relation that connects all the entities that are part of the fact.\\
The Figure~\ref{fig:sample_HE_KG} shows a pictorial representation of   two hyperedge facts given below: \\
 Fact 1: $<<PlayedTogether, Messi, Suarez, Neymar>>$ \\
 Fact 2: $<<PlayedTogether, Messi, Martínez, Di\:Maria>>$ \\
The Figure~\ref{fig:sample_HE_KG} shows a pictorial representation of   two  hyper-relational facts given below: \\
Fact 3: $<<PlayedTogether, Messi, Gerard Piqu\acute{e}>>\;(InClub: FC Barcelona),\\ (Period: 2008-21)$
\\
Fact 4: $<<PlayedTogether, Ronaldo, Gerard Piqu\acute{e}>>\;(InClub: Manchester United),\\ (Period: 2004-08)$
\\
 
 \begin{figure}[h!]
 \begin{minipage}{0.45\textwidth}
        \centering
       \includegraphics[width=1.0\linewidth]{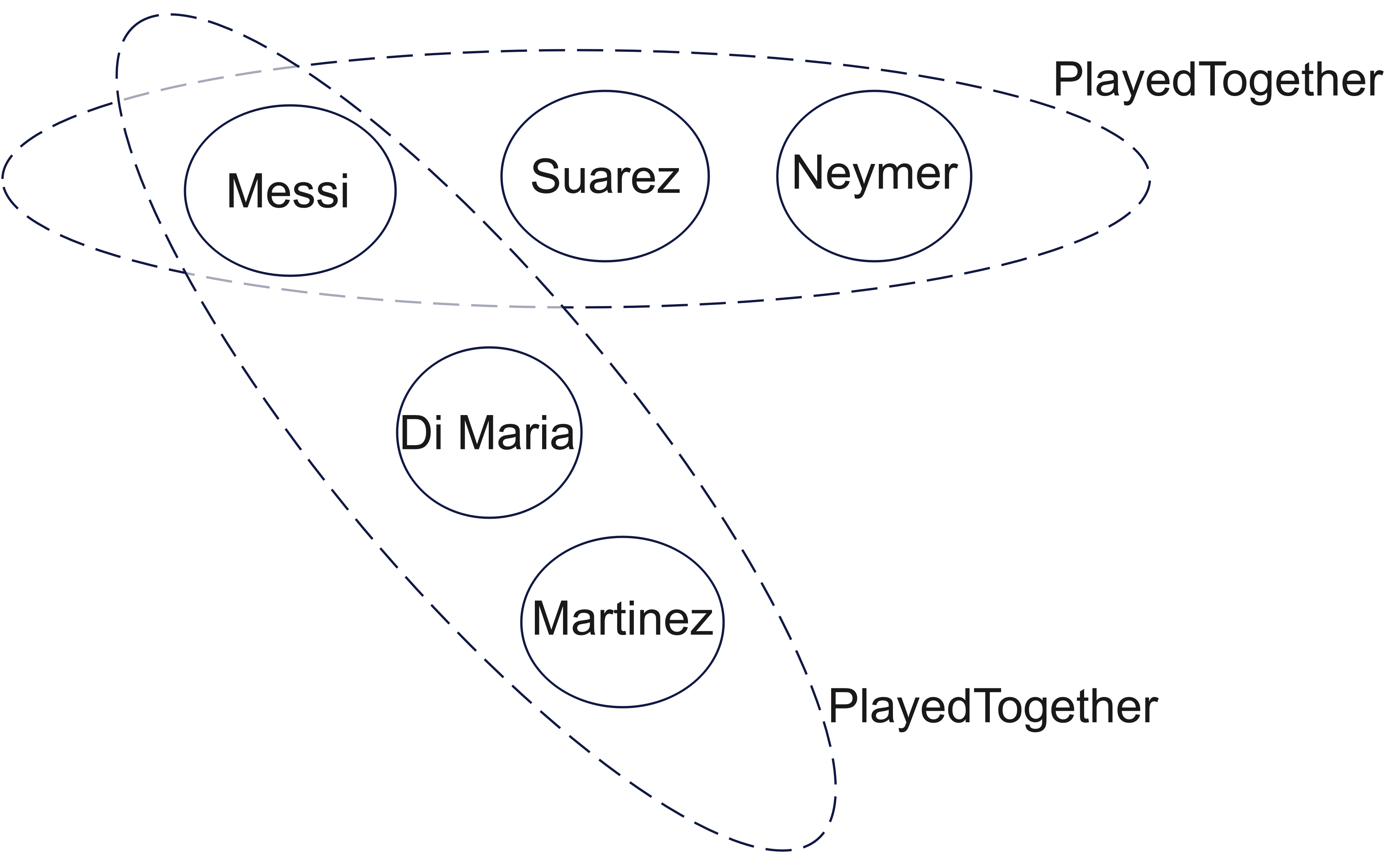}
	\caption{Representation of Hyperedge facts in KG using  Relational hyperedges}
	\label{fig:sample_HE_KG}
    \end{minipage}\hfill
    \begin{minipage}{0.45\textwidth}
        \centering
        \includegraphics[width=\linewidth]{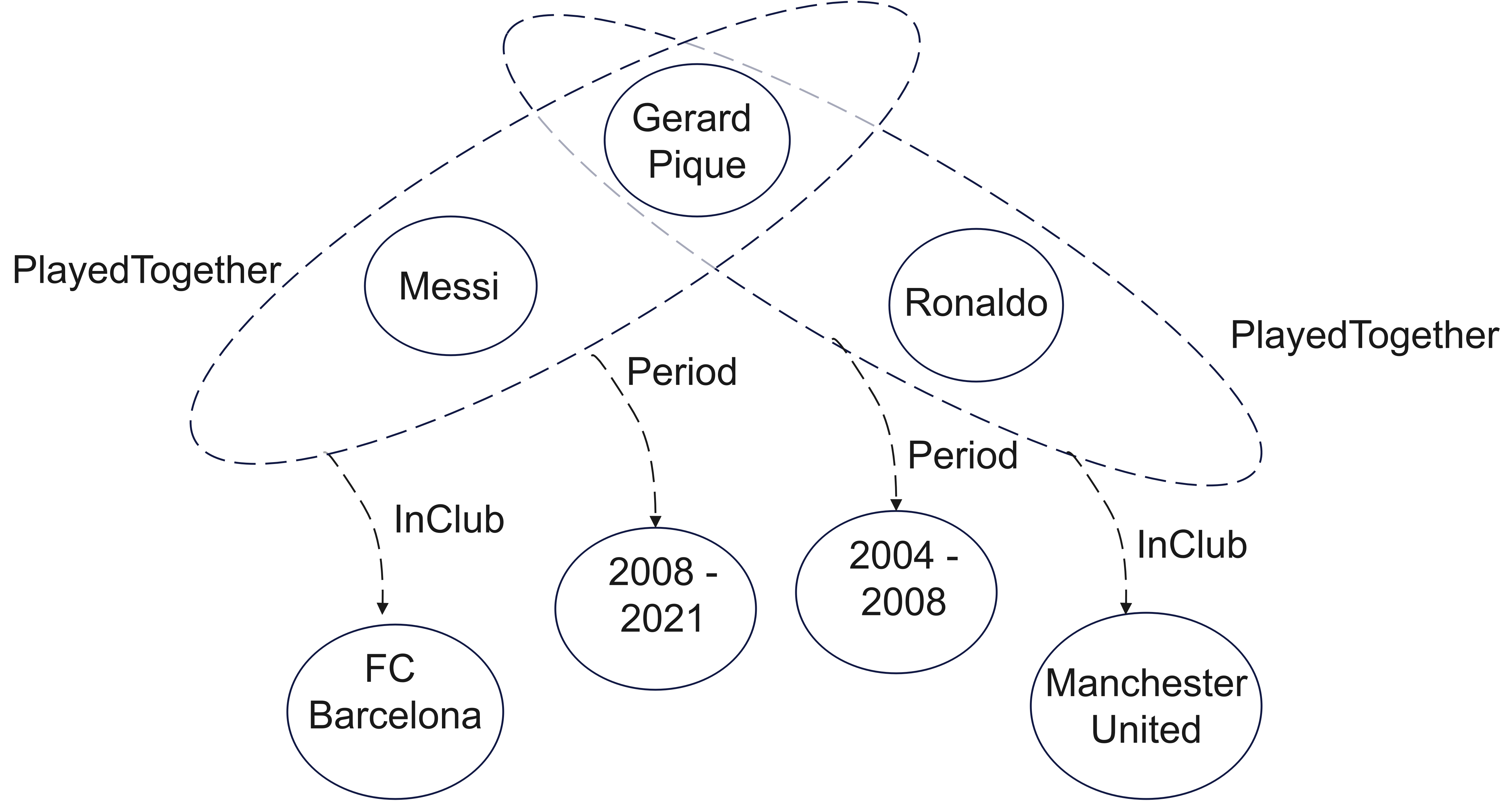}
        \caption{Representation of Hyper-relational facts in KG using Hyper-Relational edges }
        \label{fig:sample_HR_KG}
    \end{minipage}

    \label{fig:sample_HE_HR}    
\end{figure}
\vspace{-0.2in}
\textbf{Knowledge Graph Completion(KGC)} is the inferring of missing facts from the given facts in a KG. There are mainly two tasks in KGC, link prediction - inferring missing links- and node classification -  inferring missing labels. This work focuses on link prediction. Link prediction models can be transductive or inductive. \textit{Transductive} link prediction models work only in cases where all the nodes in the test dataset have been seen during training. On the other hand, \textit{Inductive} link prediction models can predict missing links even on nodes not seen during training. 

\textbf{Knowledge Graph Embedding(KGE)} is a widely adopted technique for representation learning on \kgnames. It is mainly used for KGC which is the task of inferring missing facts from the given facts - inferring missing links (link prediction) and inferring missing labels (node classification). The aim of KGE is to simplify or even make practically feasible the computations on KG.  KGE achieves this by embedding the entities and relations in a KG, which are originally in a high-dimensional space,  into a low-dimensional continuous vector space such that the similarities in the original space are preserved. A desired quality of a KGE model is high expressivity. \textit{Expressivity} refers to the ability of the model to effectively capture the structural and feature information from graphs and generate embeddings reflecting this information, allowing for more accurate inferences.

\par \textbf{GNNs} are a class of neural networks specially designed for graph structured data. Conventional neural networks, which are designed for regular structured data like images (grid-structured) and text (sequence-structured), cannot effectively handle graph data. Graphs are non-Euclidean in nature and usually have nodes with variable numbers of neighbors and without any specific ordering of neighbors. The key characteristic features of GNN architectures that make them suitable for machine learning on graphs are:

\REM{The computation graph of a GNN is based on the input graph structure itself and is leveraged for message passing, wherein information diffuses across the nodes and edges into their neighborhood. The diffusion can be controlled by means of learned parameters and the number of GNN layers.}

\begin{itemize}
  
\item Neural network structure based on the input graph: The neural network in a GNN is structured based on the input graph structure i.e. there is a neuron corresponding to each node in the graph and these neurons are linked according to the connectivity in the graph\cite{scarselli2008graph}
\item Message Passing: The feature that enables GNNs to generate representations of graphs or nodes capturing the topological, as well as the node/edge attribute information, is the information diffusion\cite{scarselli2008graph} or message passing\cite{gilmer2017neural} between the neighboring nodes. The diffusion happens across the neighbors by virtue of the special neural network structure that reflects the neighborhood connectivity as described above. 
\item Permutation invariance of neighborhood aggregation: The neighbors of nodes in a graph do not have any inherent ordering, only some arbitrary order forced by the representations like adjacency matrix. Hence the neighborhood aggregation functions in GNN are designed as permutation invariant \cite{hamilton2020graph} inorder to produce the same value irrespective of the specific arbitrary order in which the neighbors are presented in the input.    \end{itemize}
The node embedding update rule in Graph Convolutional Network(GCN)\cite{Kipf2016Semi-SupervisedNetworks}, a basic representative GNN model, is defined as below:\\

\begin{equation} \label{eq_gcn_node}
\begin{split}
h_v^{(l+1)} = \sigma\biggl(\sum\limits_{u \in N(v)}\dfrac{1}{c_{uv}}h_u^{(l)}W^{(l)}\biggr)
\end{split}
\end{equation}
where $h_v^{(l+1)}$ and $h_v^{(l)}$ are the embeddings of node v in the $(l+1)^{th}$ and $l^{th}$ GNN layers respectively,  $W^{(l)}$ is the learnable weight matrix for layer $l$, $\sigma(.)$ denotes a differentiable, non-linear activation function and $c_{uv} = \sqrt{d_vd_u}$ with $d_v$ and $d_u$ denoting the degree of node $v$ and node $u$ respectively\\
    Based on above rule, the layer-wise propagation rule for all the nodes together is written as:\\
\begin{equation} \label{eq_gcn_layer}
\begin{split}
H^{(l+1)} = \sigma\bigr(\widetilde{D}^{-\frac{1}{2}}\widetilde{A}\widetilde{D}^{-\frac{1}{2}} H^{(l)}W^{(l)}\bigl)
\end{split}
\end{equation}
where $H^{(l+1)}$ and $H^l$ are the embedding matrices of all the nodes in the $(l+1)^{th}$ and $l^{th}$ GNN layers respectively. $\widetilde{A}$ is the adjacency matrix with self connections added, $\widetilde{D}$ is the nodes degree matrix and $W^{(l)}$ is the weight matrix for layer $l$.
\section{Related works}\label{sec:Related}

\par Graph embedding based models has been the focus of research in Knowledge Graph Completion and Link prediction. The initial works catered to knowledge graphs with only binary relations. These include translational  models such as TransE\cite{BordesTranslatingData}, TransH\cite{WangKnowledgeHyperplanes} and TransR\cite{LinLearningCompletion}, Tensor factorization or semantic models such as DistMult\cite{YangEMBEDDINGBASES},  RESCAL\cite{Nickel2011AData} and SimplE \cite{KazemiSimplEGraphs}. All these models are shallow embedding models which directly optimize the embeddings for entities and relations seen during training and hence are transductive in nature. Then came neural network based models like NTN\cite{SocherReasoningCompletion} and  ConvE\cite{DettmersConvolutionalEmbeddings} which have the advantage that they use shared parameters and hence have better generalization capability and are more expressive. 
\par
Knowledge graph models supporting n-ary relations is an essential requirement for handling real-world knowledge bases. The real-world knowledge bases are generally a mix of n-ary facts that require either relational hyperedges or hyper-relational edges for representation. 
 There are  very few works on knowledge graphs with n-ary relations and, moreover, they cater to only either one of the two n-ary edge types. Works like \cite{Rosso2020BeyondPrediction} and \cite{galkin2020message} support  only hyper-relational edges whereas works like \cite{fatemi2021knowledge} 
 \cite{YadatiNeuralHypergraphs} \cite{Vazquez2023KnowledgeTaslakian} \cite{Wang2024LearningHypergraph} support only relational hyperedges.
\emph{\name  address this gap by proposing a unified model for both hyperedges and hyper-relational edges}.

\par The message passing model of Graph Neural Networks (GNN) is well suited for knowledge graph completion.
Most of the works that uses  GNN for knowledge graphs support only binary relations\cite{Kipf:2017tc}\cite{hamilton2017inductive}\cite{Velickovic:2018we} \cite{Schlichtkrull2017ModelingNetworks}. These works are based on the idea of graph convolution and uses message passing for the implementation. 
The most important related works are compared with our model \name   in Table~\ref{tab:Related-One}. The table  classifies related works based on   support for multiple relations, hyper edges, hyper relational edges. Some of these works are implemented using Graph Neural Networks.\\
\emph{Models with support for Hyper Edges} : HypE\cite{fatemi2021knowledge} proposes a decoder only model for knowledge hypergraphs. It supports hyperedges with multiple relation types and has two decoder options - HypE and HSimplE.  Both of these decoders take into consideration, the position of each entity in a relation. HSimplE uses different embeddings for an entity depending on its position in a relation whereas HypE uses a single embedding for each entity and positional convolutional weight filters for each possible position. MPNN-R\cite{YadatiNeuralHypergraphs} is a GNN-based model for multi-relational ordered and recursive hypergraphs that leverages relational message passing. HGNN\cite{feng2019hypergraph} is a GNN architecture with a propagation mechanism designed for graphs with hyper edges. The propagation is carried out in multiple steps - first from the nodes to hyperedges and then from the hyperedges to the member nodes. HGNN cater to only hypergraphs with a single relation type.
\begin{table}[h!]
        \centering
       \label{tab:compar_rel_works}
        \begin{tabular}{p{3cm}p{2cm}p{1.5cm}p{1.5cm}p{1.5cm}}
                \toprule
                {Model}& {Whether GNN-based} & \multicolumn{3}{c}{Types of Graphs supported} \\
                \cmidrule(lr){3-5}
                && GMR & GHE  & GHR \\
                \midrule
        HyPE\cite{fatemi2021knowledge} & \xmark & \cmark  & \cmark & \xmark \\
        MPNN-R\cite{YadatiNeuralHypergraphs} & \cmark & \cmark & \cmark & \xmark \\
        HINGE\cite{Rosso2020BeyondPrediction} & \xmark& \cmark  & \xmark & \cmark  \\
                STARE\cite{galkin2020message} & \cmark & \cmark & \xmark & \cmark  \\                HGNN\cite{feng2019hypergraph} & \cmark & \xmark  & \cmark & \xmark  \\
                HEHRGNN & \cmark & \cmark  & \cmark & \cmark  \\
                \bottomrule
        \end{tabular}
         \caption{Comparison of our work with related works }
(GMR:with multiple relation types, GHE:with hyperedges, GHR:with hyper-relational edges)
        \label{HEHR:tab1}
\label{tab:Related-One}
\end{table}

\emph{Models with Support for Hyper Relational Edges} : HINGE\cite{Rosso2020BeyondPrediction} is  a decoder only model for hyper-relational knowledge graphs. It scores tuples based on relatedness feature vector generated from the entity and relation embeddings using Convolutional Neural Network(CNN). STARE\cite{galkin2020message} proposes a link prediction model for Hyper-relational Knowledge Graphs comprising a GNN-based encoder and a Transformer based decoder which are jointly trained. The main contribution of this work is STaRE, the GNN-based encoder that has an additional message propagation step from the qualifier pairs attached to edges. This technique helps in capturing the qualifier context.


\emph{Other Related Works}: Semantic hyper graphs \cite{yin2025inductive}, Hypergraph convolution and hypergraph attention\cite{bai2021hypergraph} and Representation Learning on Hyper-Relational and Numeric Knowledge Graphs with Transformers\cite{chung2023representation} are some recent works that extend the above-mentioned works, but still do not cater to both types of complex facts.

\vspace{-0.2in}
\section{Motivation}
\REM{We observe that there is a gap in Knowledge graph embedding models for non-binary relational facts. Firstly of all,  research efforts in  KG modeling and KG embeddings  mostly are on binary relations.
Secondly, though there have been  work modeling non-binary facts for knowledge graph embedding, they cater to either hyperedge facts or hyper-relational facts, not both. The prior research work  pursue  each type independently.}\textbf{Why do we need a unified model supporting both hyperedges and hyper-relational edges?}
The real-world knowledge bases often have a significant share of complex n-ary facts, which cannot be represented by only  binary edges in a KG. It is observed that more than one-third of the entities in Freebase are involved in n-ary relations~\cite{wen2016representation}. More specifically, real-world knowledge bases are often a mix of complex facts that  demand both hyperedges and hyper-relational edges for their effective representation in \kgnames. The way real-world facts are modeled by humans efficiently is by using a balance of these two types of facts.

However, there are gaps in KG embedding models for n-ary relational facts - either the research efforts are mostly oriented towards binary relations or they are specific to either hyperedge facts or hyper-relational facts. The models\cite{fatemi2021knowledge}\cite{YadatiNeuralHypergraphs}  cater to only hyperedges and the models \cite{Rosso2020BeyondPrediction}\cite{galkin2020message} cater to only hyper-relational edges. Even the datasets used in the literature are designed with  a single type of these complex relations, thus failing to capture the full contextual information. The present practice is to convert real-world knowledge bases with such a mix of complex facts to simpler forms so that available models can be applied. Either all the facts are converted to binary relations and triple based models are applied, or all the facts are converted to either hyperedge or hyper-relational facts and the corresponding model is applied. The techniques commonly used for conversion of complex facts into binary relations are star-expansion (also called reification)\cite{wen2016representation}
and clique expansion  \cite{fatemi2021knowledge}. Clique expansion leads to information loss and star expansion leads to creation of specific virtual entities that prevent generalization.

Is it enough to have only either one of hyperedge or hyper-relational edge ? No. The choice of hyperedge or hyper-relational edge suitably based on the context is essential for  the proper expression of information, for setting the right perspective and for an efficient retrieval. If the  hyperedge facts in Figure~\ref{fig:sample_HE_KG} is represented   using only hyper-relational edges i.e. binary edges with qualifiers, then in order to check if Messi, Neymar, and Suarez played together at the same time in any team, we would have to compare the club names and the time periods for the three (i.e. $nC2$) binary edges (Messi-Suarez,Messi-Neymer, Neymer-Suarez). For larger groups, the retrieval becomes highly computation-intensive.
\begin{itemize}
\item A representation using only hyperedges leads to a combinatorial explosion of the number of relation types as all the qualifier attributes also have to be included as part of the primary relations .
   \REM{ \item  A hyperedge can replace hyper-relational edges by including all the required qualifier attributes but leads to a combinatorial explosion of the number of relation types}
     \item  A representation using only hyper-relational edges leads to the retrieval requiring comparison of $nC2$  binary edges (corresponding to those many entity pairs from a hyperedge of arity $n$).
\end{itemize}
 Thus, the support for both these types of n-ary facts in \kgnames\ helps in the direct representation of real-world knowledge bases,  enabling efficient retrieval and  preventing loss of information.
\par \textbf{Why Graph Neural Networks?} GNNs are proven to be powerful for KG embedding by virtue of its relational message passing that captures the topological structure of graph neighbourhoods. This is specifically important for capturing the semantic context and roles of entities and relations in KGs. GNNs combine relational message passing with the layered nonlinearities typical of neural networks. The inductive ability of GNNs by virtue of the learned transformation, aggregation and update functions makes GNNs an even more preferred choice over the transductive shallow  embedding techniques. The iterative messaging passing in GNNs lends itself to be modeled as graph parallel computations, similar to \cite{MaNeuGraph:Graphs}, making them scalable for large knowledge graphs.

    The gaps cited above regarding KG embedding models exist in GNN-based models too - they are triple oriented  or they are specific to one of the two n-ary fact types.
\REM{ HypE is a decoder only model that defines a scoring function based on positional convolutional weight filters for each possible position. The embeddings are optimized directly based on the scoring function. The HyPE scoring function is not designed to cater to hyper-relational facts which would require a two level scoring for the primary tuple and the qualifier pairs. MPNN-R is a model for multi-relational ordered and recursive hypergraphs that leverages relational message passing of GNNs. Its design os not catering to the multi-level relations of hyper-relational edges. HINGE is also a decoder only model that scores tuples based on relatedness feature vector generated from the entity and relation embeddings using Convolutional Neural Network(CNN). As the scoring function uses a fixed length relatedness feature vector, it is not designed to cater to a variable number of entities and hence cannot handle hyperedges. StaRE\cite{galkin2020message} is a encoder-decoder model for hyper-relational knowledge graphs that has a GNN-based encoder. The STARE encoder modifies the basic GNN message propagation to include the qualifiers into the edge message. HGNN is a model for general graphs with hyperedges and is designed as an extension of basic GNNs. The message propagation is modified into a two-step process by treating hyperedges as virtual entities.}To address these gaps, we propose \name, a unified embedding model for knowledge graphs with a mix of both types of n-ary facts. \name comprises a GNN-based encoder along with a unified fact representation format. Together, the encoder and the fact representation format will help to push the research towards better models for  real-world knowledge bases. 

\vspace{-0.2in}
\section{\name Unified Embedding Model  }

HEHRGNN, the proposed unified embedding model is designed to handle KGs with both hyperedges and hyper-relational edges. The two main components of the model are: i)HEHR unified fact representation format, and ii)HEHRGNN encoder. Together, these enable the model to work directly on real-world knowledge bases without having to resort to conversions or simplifications that usually lead to information loss and reduced expressivity. This in turn enables more effective downstream KG completion tasks like link prediction and node classification. The details of the HEHR unified fact representation format and the \name encoder are described in the subsections below.

\subsection{HEHR Unified fact representation format} \label{subsec:hehr_format} The unified fact representation addresses two aspects. The definition of the format in which facts are represented in KG dataset files, and the design of data structures for representing the facts in memory for embedding generation. The datasets currently used in the literature for KGs follow different formats for triple facts, hyperedge facts and hyper-relational facts. If a unified model that works across the different fact types is to be developed, a unified format is required for fact representation.  The proposed format comprises a primary tuple followed by a set of qualifier pairs. The primary tuple starts with the relation \textit{r} followed by the set of entities. The format supports a variable number of entities in the primary tuple, as well as a variable number of qualifier pairs. The examples of hyperedge facts (Fact 1 and Fact 2) and hyper-relational facts (Fact 3 and Fact 4) illustrated in section \ref{sec:background}  are written in the proposed unified fact representation format. These example facts are  also  depicted in Figure~\ref{fig:sample_HE_KG} and Figure \ref{fig:sample_HR_KG} respectively. 
\\
Two example facts having a hyperedge and a qualifier pair are given below in the unified format:\\
\unboldmath
\resizebox{\textwidth}{!}{

Fact 5: $<<PlayedTogether, Messi, Suarez, Neymar>>\;PlayedInTeam,\: FC Barcelona$ \\
}
\resizebox{\textwidth}{!}{
Fact 6: $<<PlayedTogether, Messi, Di Maria, Martinez>>\;PlayedInTeam,\: Argentina$ \\
 }
\REM{
\begin{figure}[h!]
    \centering
    \includegraphics[scale=0.30]{images/hehr_datastruct.png}
    \caption{Data structures for HEHR unified fact representation }
    \label{fig:hehr_datastruct}
\end{figure}
}
\begin{figure}[h!]
\centering

\begin{minipage}{0.3\textwidth}
\centering
\caption*{Entity Map}
\begin{tabular}{|c|l|}
\hline
\textbf{ID} & \textbf{Entity} \\ \hline
0 & Messi \\ \hline
1 & Suarez \\ \hline
2 & Neymar \\ \hline
3 & Di Maria \\ \hline
4 & Martinez \\ \hline
5 & Argentina \\ \hline
6 & FC Barcelona \\ \hline
7 & --- \\ \hline
\end{tabular}
\end{minipage}
\hfill
\begin{minipage}{0.3\textwidth}
\centering
\caption*{Relation Map}
\begin{tabular}{|c|l|}
\hline
\textbf{ID} & \textbf{Relation} \\ \hline
0 & BornIn \\ \hline
1 & PlayedTogether \\ \hline
2 & PlayedInTeam \\ \hline
3 & --- \\ \hline
4 & --- \\ \hline
\end{tabular}
\end{minipage}
\hfill
\begin{minipage}{0.3\textwidth}
\centering
\caption*{HyperEdge Relation Type Map}
\begin{tabular}{|c|c|}
\hline
\textbf{\shortstack{HyperEdge\\ Index}} & \textbf{\shortstack{Relation\\ ID}} \\ \hline
0 & 1 \\ \hline
1 & 1 \\ \hline
2 & --- \\ \hline
3 & --- \\ \hline
4 & --- \\ \hline
\end{tabular}
\end{minipage}

\vspace{1cm}
\begin{minipage}{0.4\textwidth}
\centering
 \caption*{HyperEdge Entity Details}
\begin{tabular}{|c|c|c|c|c|c|c|}
\hline
\textbf{HyperEdge Index} & 0 & 0 &0&1 & 1 & 1 \\ \hline
\textbf{Entity ID}       & 0 & 1 & 2 & 0 &3&4\\ \hline
\end{tabular}
\end{minipage}%
\hfill
\begin{minipage}{0.3\textwidth}
\centering
\caption*{HyperEdge Qualifier Details}
\begin{tabular}{|c|c|c|c|}
\hline
\textbf{HyperEdge Index} & 0 & 1 & --- \\ \hline
\textbf{Qual. Relation}     & 2 & 2 & --- \\ \hline
\textbf{Qual. Entity}    & 5 & 6 & --- \\ \hline
\end{tabular}
\end{minipage}
\caption{Data structures for HEHR unified fact representation }
    \label{fig:hehr_datastruct}
\end{figure}

The data structures used by \name\ for unified fact representation are illustrated  in Figure \ref{fig:hehr_datastruct} by using Fact 5 and Fact 6.  Currently, the design assumes two levels- a primary tuple and a set of qualifier pairs - with the qualifiers restricted to one relation and one entity. The format can be easily extended to support multiple recursive levels of information and a single qualifier with multiple entities. 
\subsection{HEHRGNN encoder}
The core of the unified embedding model is the HEHRGNN encoder that generates entity and relation embeddings from input KG data represented in the HEHR unified format. The entity and relation embeddings can be used for downstream tasks like link prediction and node classification. The details of the HEHRGNN encoder design and the message propagation algorithm are described below.  

\vspace{-0.1in}
\subsubsection{Design of the HEHRGNN encoder}
The HEHRGNN encoder is designed with a novel GNN message propagation scheme capable of aggregating information from neighborhoods connected via hyperedges as well as hyper-relational edges.\REM{ In non-GNN models for KG embedding, the individual tuples in the training sample are used independently one after the other for optimizing the embeddings without considering the whole graph structure.}HEHRGNN, like other GNN-based models, initially constructs the overall graph structure from all the facts. Thereafter, message propagation over the graph neighborhood helps to capture the local and global graph structure for generating the embeddings.

In HEHERGNN, all the entities, whether appearing in primary tuples or in qualifier pairs are treated as the same class of entities.  However, the information passed from each node is aggregated with suitable learned weights based on their role in a particular fact instance. 
This makes the model capable of generating embeddings that capture the multiple  semantic contexts or roles of each entity more effectively. HEHRGNN encoder design supports multiple GNN layers in order to be able to capture multi-hop neighbourhood contexts.  
The evaluation of the generated embeddings using link prediction task shows model effectiveness for combined  hyperedge and hyper-relational datasets. It also shows improved performance of HEHRGNN over prior independent models for hyperedges \cite{fatemi2021knowledge} and hyper-relational edges\cite{galkin2020message}. 

\subsubsection{HEHRGNN Message Propagation Algorithm}

\begin{figure}[h!]
    \begin{tikzpicture}[
    node distance=8mm and 14mm,
    every node/.style={font=\small},
    state/.style={circle, draw, minimum size=8mm},
    hidden/.style={ellipse, draw, dashed, minimum width=12mm, minimum height=28mm},
    arrowG/.style={->, thick, green!70!black},
    arrowB/.style={->, thick, blue!60},
    arrowR/.style={->, thick, red!70}
]


\node[state] (D) at (0,3) {D};
\node[state] (E) at (0,2) {E};
\node[state] (F) at (0,1) {F};

\node[state] (C) at (0,-0.5) {C};
\node[state] (B) at (0,-1.5) {B};
\node[state] (A) at (0,-2.5) {A};

\node[state] (H) at (0,-4) {H};

\node[hidden] (e2l) at (3,1.8) {e2};
\node[hidden] (e1l) at (3,-2) {e1};

\draw[arrowG] (D) -- (e2l);
\draw[arrowG] (E) -- (e2l);
\draw[arrowG] (F) -- (e2l);

\draw[arrowG] (C) -- (e1l);
\draw[arrowG] (B) -- (e1l);
\draw[arrowG] (A) -- (e1l);

\draw[arrowB] (F) -- (e1l);
\draw[arrowB] (B) -- (e2l);
\draw[arrowB] (H) -- (e1l);
\node[below=4mm of H] {\shortstack{Gather from Primary \\and Qualifier Nodes}};
\draw[very thick] (4,-4.8) -- (4,4);


\node[hidden] (e2m) at (5.5,1.8) {e2};
\node[hidden] (e1m) at (5.5,-2) {e1};

\node[draw, rectangle, rounded corners, minimum width=14mm, minimum height=7mm]
(R1) at (7.5,-0.1) {R1};

\draw[arrowR] (e2m.east) -- (R1.west);
\draw[arrowR] (e1m.east) -- (R1.west);
\node[below =32mm of e1m.east] {Apply on Relation Types};
\draw[very thick] (8.5,-4.8) -- (8.5,4);


\node[hidden] (e2r) at (9.5,1.8) {e2};
\node[hidden] (e1r) at (9.5,-2) {e1};

\node[state] (Dr) at (11.5,3) {D};
\node[state] (Er) at (11.5,2) {E};
\node[state] (Fr) at (11.5,1) {F};

\node[state] (Br) at (11.5,-0.8) {B};
\node[state] (Ar) at (11.5,-1.8) {A};
\node[state] (Cr) at (11.5,-2.8) {C};
\node[state] (Hr) at (11.5,-4) {H};

\draw[arrowG] (e2r) -- (Dr);
\draw[arrowG] (e2r) -- (Er);
\draw[arrowG] (e2r) -- (Fr);

\draw[arrowG] (e1r) -- (Br);
\draw[arrowG] (e1r) -- (Ar);
\draw[arrowG] (e1r) -- (Cr);

\draw[arrowB] (e2r) -- (Br);
\draw[arrowB] (e1r) -- (Fr);
\draw[arrowB] (e1r) -- (Hr);
\node[below=24mm   of e1r.east] {\shortstack{Scatter  to Primary \\and Qualifier Nodes}};
\end{tikzpicture}
   \caption{HEHRGNN message propagation}
    \label{fig:msg_propgn}
\end{figure}
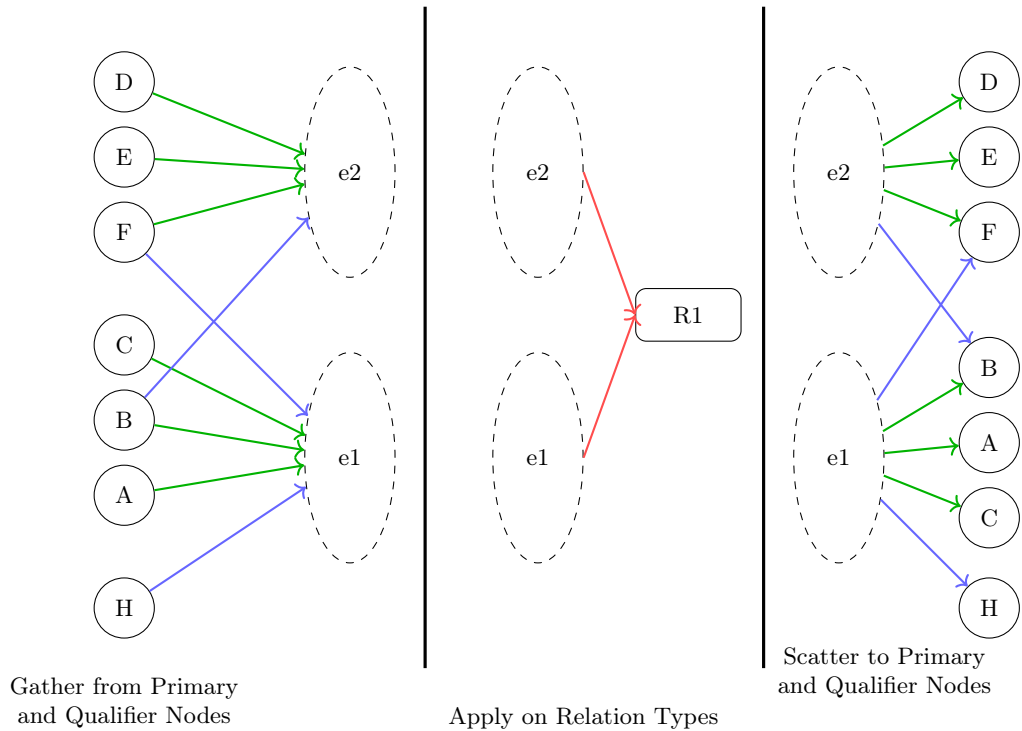

The HEHRGNN message propagation can be broken down as given below into a sequence of Gather, Apply and Scatter based on a hyperedge centric view:
\begin{enumerate}
    \item Gather from the primary and qualifier nodes of hyperedges and update the hyperedge embeddings.
    \item Apply the updated embeddings of hyperedge instances to the corresponding relation types  and update the relation embeddings. 
    \item Scatter the updated hyperedge embeddings to the primary and qualifier nodes, and update the node embeddings.
\end{enumerate}
 Here, the member nodes of the hyperedges are referred to as \textit{primary nodes} and the nodes in the qualifier pairs are called \textit{qualifier nodes}. The separation of the message propagation into the steps as given above clearly brings out the main information flows in the knowledge graph that help in the generation of embeddings. 
 An overview of the message propagation steps is depicted in Figure \ref{fig:msg_propgn}. As can be seen from the figure, the design of  message propagation in HEHRGNN uses a hyperedge centric view, different from the vertex-centric view commonly used in \cite{MaNeuGraph:Graphs} and other GNN models. This is done order to support hyperedge and hyper-relational facts with variable number of primary nodes and qualifier pairs. 

\unboldmath

\begin{algorithm}[h!]
\caption{HEHRGNN Encoder Message propagation algorithm for Embedding generation}
\label{alg:alg_encoder}
\begin{algorithmic}[1]
\algrenewcommand\algorithmicrequire{\textbf{Input:}}
\algrenewcommand\algorithmicensure{\textbf{Output:}}
\Require Set of facts $\mathcal{F}$ corresponding to a Knowledge Graph $\mathcal{KG}$ in the HEHRGNN unified format; 

\Ensure Embedding vector $h_v$ for all entities $v \in \mathcal{V}$; Embedding vector $h_r$ for all relations $r \in \mathcal{R}$; 

\State From $\mathcal{F}$, derive the set of entities $\mathcal{V}$, the set of of relations $\mathcal{R}$, set of primary hyperedges $\mathcal{E}$ and the set of  qualifier pairs $\mathcal{Q}$ 
\State Construct the graph structure using $\mathcal{V}$, $\mathcal{R}$, $\mathcal{E}$ and $\mathcal{Q}$
\State $h^0_v \leftarrow x_v, \forall v \in \mathcal{V}$; \label{alg:ent_init}
\State $h^0_r \leftarrow x_r, \forall r \in \mathcal{R}$; \label{alg:rel_init}
\For{\texttt{$l=1...L$}}    \label{alg:for_loop_num_layers}
   \For{\texttt{$e \in \mathcal{E}$}} \label{alg:for_loop_hyperedges}
\State $h^l_{P\_nodes(e)} \leftarrow {Gather\_from\_Primary}(\{h^{l - 1}_u, \forall u \in P\_nodes(e)\});$ \label{alg:msg_ptoedge}
\State $h^l_{Q\_nodes(e)} \leftarrow {Gather\_from\_Qual}(\{h^{l - 1}_u, \forall u \in Q\_nodes(e)\});$ \label{alg:msg_qtoedge}
\State $h^l_{e} \leftarrow {Apply\_Hyperedge\_Weighted}(h^{l-1}_e, h^{l}_{P\_nodes(e)},h^l_{Q\_nodes(e))})$; \label{alg:edge_combine}
    \EndFor
\State
\State $h^k_r \leftarrow Apply\_Relation(h^{l-1}_r, \{h^l_e, \forall e \in \mathcal{E} for\;which\; edge\; type=r\})$; \label{alg:rel_embed}
\State    
    \For{\texttt{$v \in \mathcal{V}$}}\label{alg:for_loop_nodes}
\State $h^l_{P\_edges(v)} \leftarrow {Scatter\_to\_Primary}(\{h^l_e, \forall e \in P\_edges(v)\});$ \label{alg:edge_top}
\State $h^l_{Q\_edges(v)} \leftarrow {Scatter\_to\_Qualifier}(\{h^l_e, \forall e \in Q\_edges(v)\});$ \label{alg:edge_toq}
\State $h^l_{v} \leftarrow {Apply\_Entities\_Weighted}(h^{l-1}_v, h^{l}_{P\_edges(v)},h^l_{Q\_edges(v))})$; \label{alg:node_combine}
    \EndFor
\EndFor 
\end{algorithmic}
\end{algorithm}

\boldmath

The detailed pseudocode for the propagation algorithm is given in Algorithm~\ref{alg:alg_encoder}.
The Lines~\ref{alg:ent_init} and ~\ref{alg:rel_init} initializes the entity and relation embeddings respectively. The outermost \textit{for} loop starting at   Line~\ref{alg:for_loop_num_layers} corresponds to the iteration over multiple GNN layers. $L$ is the configured number of GNN layers. The \textit{for} loop starting at Line \ref{alg:for_loop_hyperedges} corresponds to the \textit{Gather} phase of the message propagation steps. The loop iterates over all hyperedges $e$ to \texttt{gather} messages from their \textit{primary-nodes} (P\_nodes in Line~\ref{alg:msg_ptoedge}) and \textit{qualifier-nodes} (Q\_nodes in Line~\ref{alg:msg_qtoedge}) and then \textit{aggregate} them and \textit{apply} to the hyperedges (Line~\ref{alg:edge_combine}). 
 The \textit{apply to hyperedge}in Line~\ref{alg:edge_combine} is implemented as a weighted average of the aggregated messages from \textit{primary nodes} and \textit{qualifier nodes}.  The Line~\ref{alg:rel_embed} corresponds to  the \textit{Apply} phase on relations, that apply the  updated edge embeddings onto the relation types and update the relation embeddings.
 \REM{ Each hyperedge embedding contributes to update the embedding of its relation type. Or in other words}The embedding of each relation type $r\in R$ is impacted by all the hyperedge instances having  relation type $r$.   The \texttt{for} loop starting in Line~\ref{alg:for_loop_nodes} corresponds to the \textit{scatter} phase. The loop iterates over all the entities to collect and apply the messages scattered by the hyperedges. Each node  gets message contributions from all hyperedges in which it is a \textit{primary node} (Line\ref{alg:edge_top}) or is a  \textit{qualifier node} (Line~\ref{alg:edge_toq}). The apply node (Line~\ref{alg:node_combine}) function is implemented as a weighted average of  messages, with different weights based on the node role: \textit{primary node} or a \textit{qualifier node}.  

As the \textit{gather} and \textit{apply} steps for all the hyperedges can be carried out in parallel, the \texttt{for} loop in Line~\ref{alg:for_loop_hyperedges} is implemented in parallel. Similarly the \texttt{for} loop over nodes is implemented in parallel. Moreover, the first two steps in both these \textit{for loops}  are executed in parallel.

The weights used in the GNN layer functions are the major learnable parameters of the HEHRGNN encoder that contribute to the generation of embeddings. The GNN layer functions that form part of the design are the Gather, Apply and Scatter functions that appear in Algorithm~\ref{alg:alg_encoder}. An important hyper-parameter of the encoder that is configured or tuned empirically  is $L$, the number of GNN layers.
\subsubsection{Discussion on the HEHRGNN update rule}
Here we bring out the difference in the update rule used in HEHRGNN with respect to the update rule used in GCN\cite{Kipf2016Semi-SupervisedNetworks} that is mentioned above in equation \ref{eq_gcn_node}. It can be seen that GCN does not learn relation embeddings and has update rule only for the entities. HEHRGNN learns embeddings for both entities and relations. As explained before, a relation embedding is updated by aggregating contributions from all hyperedge instances of that relation type. 
The update rules of HEHRGNN for hyperedges, relations and entities are given below:\\

 For hyperedge e,\\
 \unboldmath
\begin{equation} \label{eq_hyperedge}
    \resizebox{\textwidth}{!}{
$h_e^{(l+1)} = \sigma\biggl(W_{PN} \dfrac{1}{|P\_nodes(e)|}\sum\limits_{u \in P\_nodes(e)} h_u^{(l)}\biggr) + \sigma\biggl(W_{QN} \dfrac{1}{|Q\_nodes(e)|}\sum\limits_{u \in Q\_nodes(e)} h_u^{(l)}\biggr)$
}
\end{equation}

For relation r,\\ 
 \begin{equation} \label{eq_reln}
h_r^{(l+1)} = \sigma\biggl(W_R \dfrac{1}{|instances(r)|} \sum\limits_{ \forall e \in \mathcal{E}\; for\;which\; edge\; type=r\}} h_e^{(l+1)}\biggr)
\end{equation}
For entity v,\\
 \unboldmath
 \begin{equation} \label{eq_entity}
     \resizebox{\textwidth}{!}{
$h_v^{(l+1)} = \sigma\biggl(W_{PE} \dfrac{1}{|P\_edges(v)|}\sum\limits_{e \in P\_edges(v)} h_e^{(l+1)}\biggr) + \sigma\biggl(W_{QE} \dfrac{1}{|Q\_edges(v)|}\sum\limits_{e \in Q\_edges(e)} h_e^{(l+1)}\biggr)$
}
\end{equation}

\subsection{Discussion on the benefits of the unified embedding model}
 HEHRGNN model can directly process real-world datasets containing a mix of both types of n-ary facts without requiring any conversion. This capability enables the model to capture the semantic roles of entities, their contexts, and connections present in the real-world datasets without any information loss.
 
 The HEHRGNN model is designed to directly process real-world datasets containing a mix of n-ary facts without requiring any conversion. The specific design elements of the model supporting this capability include the unified fact representation format, data structures for storing and processing the n-ary facts, and the novel GNN message propagation mechanism. By virtue of this design, the proposed model gets to see the realistic roles, contexts, and perspectives of each entity and relation, exactly as present in the knowledge base. This means the embeddings generated by the model capture the semantic roles of all entities and relations in a set of facts without any loss of information or loss of semantic perspective. 
 
The following examples illustrate how each type of fact carries a distinct perspective in the real-world context. \\~\\Examples of hyperedge facts: \\
     Fact 6: $<<EducationalQualification, Albert Einstein, ETH Zurich,\\ Bachelor Degree,Mathematics Major, 1900>>$ \\
     Fact 7: $<<EducationalQualification, Albert Einstein, University of Zurich,\\ Ph.D. Degree, Physics,1905>>$ \\~\\
Examples of hyper-relational facts: \\
     Fact 8: $<<GraduatedFrom, Albert Einstein, ETH Zurich>>$ \\$AcademicDegree, Bachelor, AcademicMajor, Mathematics, Year, 1900$ \\
     Fact 9: $<<GraduatedFrom, Albert Einstein, University of Zurich>>$ \\$AcademicDegree,PhD, AcademicMajor, Physics,Year,1905$ \\~\\
The hyperedge representation(Fact 6 and Fact 7) is more suited when we are interested in the educational qualifications of Albert Einstein with all the details, whereas the hyper-relational representation(Fact 8 and Fact 9) is more suited when we are interested in the institutes Albert Einstein attended, with the qualifiers being optional but useful information.  
HEHRGNN model enables the representation of facts in whichever format is appropriate for conveying the right perspective of information, and facilitates the processing and capturing of this information for embedding generation.

\REM{\todo{Whether this portion is required---->!!!!! Following the expressivity theorem in \cite{fatemi2021knowledge}, we can conjecture that HEHRGNN can generate embedding vector assignments for each of the entities($\mathcal{E}$) and relations($\mathcal{R}$ ) in a knowledge base $\mathcal{F}$ such that the decoder can score positive facts higher. This theorem assumes that the embedding dimension is chosen appropriately based on the number of ground truth facts and the maximum arity.
HEHRGNN fits the definition of Hypergraph Relational Message Passing Neural Networks (HR-MPNNs) as defined in \cite{Huang2025LinkHypergraphs}. As the equivalence of  HR-MPNNs and the hypergraph relational 1-WL test, a variant of the 1-dimensional Weisfeiler-Leman test is shown in \cite{Huang2025LinkHypergraphs},  HEHRGNN also has the same expressive power. }}

\subsection{Implementation Details} \label{sec:implement}


An end-to-end encoder-decoder model for link prediction on KGs is implemented along with the training module to evaluate the proposed HEHRGNN embedding model. The implemented decoder is for the link prediction task and has two scoring function options - HypE and modified DistMult, both proposed in \cite{fatemi2021knowledge}. The implementation is done using PyTorch, with the message propagation implemented by customising the  \textit{MessagePassing} class, the base class for convolutional layers in the PyTorch Geometric library. The source code is publicly available on github \footnote{\url{https://github.com/gitrajeshr/HEHRGNN}}.

The implementation uses tensor-based data structures, as per the design detailed in section \ref{subsec:hehr_format}, for storing and processing the \kgname data that is fed in the unified format. The primary tuples and the associated qualifier pairs are stored as Coordinate lists(COO) in separate tensors and are linked together by edge indices. This design enables handling a variable number of entities in the primary tuple and a variable number of qualifier pairs. The two main operations involved in the embedding generation in HEHRGNN are the node and edge centric propagations and the scatter-reduce used for updating the relation embeddings from edge instances. As both these operations are parallelized by the PyTorch backend, HEHRGNN is able to leverage the compute resources efficiently for embedding generation.

\subsubsection{Model Training}
The training involves message propagation across the entire graph to generate embeddings, and then computing the loss with respect to  predictions for the samples in the batch using the generated embeddings. HEHRGNN follows the negative sample strategy used in \cite{fatemi2021knowledge} and  \cite{BordesTranslatingData} for training and evaluation. A set of negative samples are generated for each positive sample in the batch by replacing each entity in the tuple, one at a time, with random entities. The number of negative samples generated for each positive sample is $N \times |r|$ where $N$ is the negative samples ratio configured as a hyperparameter of the model and $|r|$ is the arity of the positive sample. $N$ is configured as 10 in the reported experiments.     
The loss function and the optimizer used for learning and optimization of the parameters are Binary Cross Entropy Loss(BCEL) and Adam Optimizer respectively.  The loss is computed from the scores produced by the scoring function for each of the samples, including positive and negative, in the batch. The scores are normalized using the sigmoid function. The link prediction performance of the model is evaluated using rank based evaluation metrics. These aspects are further detailed in the experimental evaluation section.

\vspace{-0.23in}
 \section{Experimental Evaluation}
The proposed HEHRGNN encoder is experimentally evaluated using the PyTorch-based encoder-decoder model implementation. The encoder-decoder model is trained  to generate embeddings and carry out link prediction. The performance of the encoder on multiple datasets is assessed by measuring link prediction performance using a set of rank-based evaluation metrics. The experimental settings used and various analyses with respect to the experimental results are detailed in the subsections below.
\subsection{Experimental settings}
The details of the hardware and software platforms, the datasets and the evaluation metrics used for the experiments are described below.\\
 \par\textbf{Hardware and Software platform:} The results presented and discussed in the paper are with respect to the experiments run on Nvidia H100 GPU with 94GB RAM. This platform is able to accommodate even the larger datasets and thereby maintain uniform settings across all the experiments. The software platform used for implementation and experiments is PyTorch with PyTorch Geometric library and CUDA backend.\\
 
\textbf{Datasets:} Experiments on link prediction have been run with multiple datasets including those used and referred in the related works on knowledge hypergraphs and hyper-relational knowledge graphs. The details of datasets used are listed in  Table~\ref{tab:datasets}.

\REM{
\begin{itemize}
\item JF17K: A hyperedge dataset extracted and filtered from Freebase, by \cite{wen2016representation} for research on embedding knowledge bases with n-ary relations. 
\item FB-AUTO and M-FB15K: Two hyperedge datasets created from Freebase by \cite{fatemi2021knowledge}. 
\item WD50K : A hyper-relational dataset derived from Wikidata statements (Wikidata RDF dump) by \cite{galkin2020message}. 
\item WikiPeople: A hyper-relational dataset extracted from Wikidata, by\cite{guan2019link},  that focuses on entities of type human. 
\item Combined-HEHR dataset: For evaluating the effectiveness of HEHRGNN on datasets with a mix of hyperedge and hyper-relational facts, we  combined WD50K and JF17K into one, which is named as Combined-HEHR (Hyperedge Hyper-relational) dataset.
\end{itemize}
 The statistics of these datasets are given in Table \ref{tab:datasets}.\\}
 \begin{table}[h!]
	\centering
    	 \caption{Dataset Details (HE: Hyperedge, HR: Hyper-relational) }
	\label{tab:datasets}
\addtolength{\tabcolsep}{-5pt}
	\begin{tabular}{ccccccc}
		\toprule
		 Dataset & Type & Train facts & Valid facts & Test facts & $|Entities|$ & $|Relations|$\\
		\midrule
           JF17K\cite{wen2016representation} & HE & 61911 & 15822 & 24915 & 28,645 & 322
		\\
        FB-AUTO\cite{fatemi2021knowledge} & HE & 6778 & 2255 & 2180 & 3410 & 8 \\
         M-FB15K\cite{fatemi2021knowledge} & HE & 415375 & 39348 & 38797 & 10314 & 71
		\\         
		WD50K\cite{galkin2020message} & HR & 166,435 & 23,913 & 46,159 & 47,156 & 532 \\              WikiPeople\cite{guan2019link} & HR & 305725 & 38223 & 38281 & 47766 & 305 \\
        Combined-HEHR & HEHR & 228346 & 24913 & 71074 & 75801 & 854\\
    	\bottomrule
	\end{tabular}
   \end{table}
   
\textbf{Evaluation Metrics Used:} We have used the rank-based evaluation metrics to rank the positive triple in a sorted list of triple scores.Hits@k($H_k$ for k=1,3,5,10) and Mean Reciprocal Rank(MRR)\cite{Hoyt2022AGraphs} are the rank-based metrics commonly reported in the literature. These are aggregate statistics for a set of predictions.\\ \textit{Hits@k($H_k$)} indicates the fraction of positive triples among all the predictions that appeared within k ranks in the sorted list of scores. The higher the value of Hits@k, the better the prediction performance. It is computed as:\\  $H_k(r_1,r_2,....r_n) = \dfrac{1}{n} \sum\limits_{i=1}^n\mathbf{I}[r_i<=k]\qquad \in [0,1]$\\
\textit{Mean Reciprocal Rank(MRR)} indicates the arithmetic mean of the reciprocal of ranks of the positive triples among all the predictions. The higher the value of MRR, the better the prediction performance. It is computed as:\\ $MRR(r_1,r_2,....r_n) = \dfrac{1}{n} \sum\limits_{i=1}^n \dfrac{1}{r_i}\qquad \in [0,1]$

\vspace{-0.13in}
\subsection{Discussion of  Link Prediction Results}

Here we present the results of link prediction experiments with the proposed model on multiple dataset types - Hyperedge, Hyper-Relational and combined. The performance values of the baseline models of the corresponding dataset types are also given to enable a comparison of the HEHRGNN performance. 

The results of link prediction experiments with the proposed model on hyperedge and hyper-relational datasets are presented in Table \ref{tab:lp_results_he}  and Table \ref{tab:lp_results_hr} respectively, along with values for the corresponding baseline models. The values for the baseline models are taken from \cite{fatemi2021knowledge} for hyperedge facts and from \cite{galkin2020message} for hyper-relational facts. The best values are shown in bold. The results indicate that the proposed model is able to improve upon the link prediction performance of the baseline models except for the FB-AUTO dataset. We conclude that FB-AUTO, being a very small dataset,  is not sufficient for training the GNN-based encoder. The link prediction results for STARE model reported in \cite{galkin2020message} are the average of subject and object predictions done separately. As STARE performs link prediction as a sequence prediction - of the \textit{(subject, relation, object)}  sequence - using a transformer-based decoder, it needs to do the object and subject predictions separately. However, as HEHRGNN uses the HypE decoder, that generates a score for the potential tuples based on the entity and relation embeddings and a position-specific weight, it need not distinguish between object and subject predictions. Hence, HEHRGNN is not required to add inverse relations during the training, as is done by STARE.

\begin{table}[h!]
	\centering	
    \footnotesize
      \caption{Link Prediction Results for Hyperedge Facts. Results of r-SimplE, m-DistMult, m-CP, m-TransH, HSimplE and HyPE are taken from \cite{fatemi2021knowledge}}
	\label{tab:lp_results_he}
    \newrobustcmd{\B}{\bfseries}
\addtolength{\tabcolsep}{-2pt}
\resizebox{\textwidth}{!}{
\begin{tabular}{ccccccccccccc}
		\toprule
		 Model &  \multicolumn{4}{c} {JF17K} & \multicolumn{4}{c} {FB-AUTO} & \multicolumn{4}{c} {M-FB15K}\\
         \cmidrule(lr){2-5} \cmidrule(lr) {6-9}
         \cmidrule(lr) {10-13}
         &MRR & Hits@1 & Hits@3 & Hits@10 &MRR & Hits@1 & Hits@3 & Hits@10&MRR & Hits@1 & Hits@3 & Hits@10\\
		\midrule
        r-SimplE & 0.102 & 0.069 & 0.112 & 0.168 & 0.106 & 0.082 & 0.115 & 0.147 & 0.051 & 0.042 & 0.054 & 0.070 \\
        m-DistMult & 0.463 & 0.372 & 0.510 & 0.634 & 0.784 & 0.745 & 0.815 & 0.845 & 0.705 & 0.633 & 0.740 & 0.844 \\
        m-CP & 0.391 & 0.298 & 0.443 & 0.563 & 0.752 & 0.704 & 0.785 & 0.837 & 0.680 & 0.605 & 0.715 & 0.828 \\
        m-TransH & 0.444 & 0.370 & 0.475 & 0.581 & 0.728 & 0.727 & 0.728 & 0.728 & 0.623 & 0.531 & 0.669 & 0.809 \\
        HSimplE & 0.472 & 0.378 & 0.520 & 0.645 & 0.798 & 0.766 & 0.821 & 0.855 & 0.730 & 0.664 & 0.763 & 0.859 \\
        	 HypE &  0.494  & 0.408 & 0.538 & 0.656 & \B 0.804& \B 0.774 & \B 0.823 &  \B 0.856 & 0.777 & 0.725 & 0.800 & 0.881  \\
              HEHRGNN &  \B 0.712  & \B 0.596 & \B 0.752 & \B 0.865 & 0.711 & 0.562 & 0.826 & 0.877 & \B 0.810 & \B 0.789 & \B 0.830 & \B 0.910  \\       
		\bottomrule
	\end{tabular}
  }
\end{table}

\begin{table}[h!]
	\centering	
    \footnotesize
      \caption{\shortstack{Link Prediction Results on Hyper-Relational Facts. Results of m-TransH,\\ RAE, NaLP-Fix, HINGE, STARE are taken from \cite{galkin2020message}}}\label{tab:lp_results_hr}
    \newrobustcmd{\B}{\bfseries}
\addtolength{\tabcolsep}{-5pt}
\begin{tabular}{ccccccccc}
		\toprule
		 Model &  \multicolumn{4}{c} {WD50K} & \multicolumn{4}{c} {WikiPeople} \\
         \cmidrule(lr){2-5} \cmidrule(lr) {6-9}
          &MRR & Hits@1 & Hits@3 & Hits@10 &MRR & Hits@1 & Hits@3 & Hits@10\\
		\midrule
      m-TransH & - & - & - & - &  0.063 & 0.063 & - & 0.300 \\
RAE & - & - & - & - &           0.059 & 0.059 & - & 0.306 \\
NaLP-Fix & 0.177 & 0.131 & - & 0.264 &   0.420 & 0.343 & - & 0.556 \\
HINGE & 0.243 & 0.176 & - & 0.377 &  0.476 & 0.415 & - & 0.585 \\
STARE & 0.349 & 0.271 & - & 0.496 & 0.491 & 0.398 & 0.592 \\
HEHRGNN & \B 0.759  & \B 0.584 & \B 0.743 & \B 0.925 & \B 0.752 &\B 0.712 &\B  0.810 &\B 0.905 \\
		\bottomrule
	\end{tabular}
  
\end{table}
 The experiments with combined HEHR dataset shows that the proposed model is able to produce entity and relation embeddings from a mix of hyperedge and hyper-relational facts with a reasonable link prediction performance.  As there are no prior works, to the best of our knowledge, that presents results on a combined HEHR dataset, we do not have any baselines to compare. However, as presented in Table \ref{tab:lp_results_comb}, we show that the results are  comparable to those of separate models for the two fact types.

 \vspace{-0.2in}
  \begin{table}[h!]
        \centering
\begin{minipage}{0.5\textwidth}
        \centering
          \caption{\shortstack{Link Prediction Results on Mix of \\ Hyperedge and Hyper-Relational Facts}}\label{tab:lp_results_comb}
\addtolength{\tabcolsep}{-5pt}
 
        \begin{tabular}{ccccc}
                \toprule
        Model &  \multicolumn{4}{c} {Combined-HEHR dataset}\\
         \cmidrule(lr){2-5}
                   &  MRR & Hits@1 & Hits@3 & Hits@10\\
                \midrule
                 HEHRGNN &  0.612  & 0.385 & 0.458 & 0.562  \\
                \bottomrule
        \end{tabular}       
\end{minipage}
\end{table}
\vspace{0.5in}

\begin{figure}[h]
    \centering
\begin{tikzpicture}
  \begin{axis}[
    xlabel={Embedding Dimension},
    ylabel={Link Prediction Performance},
    legend style={at={(1.05,1)}, anchor=north west},
    grid=both,
    scaled ticks=false,
    xtick={100,128,200,256, 300},
    yticklabel style={/pgf/number format/.cd,fixed,precision=3},]

\addplot[
    color=green,
    thick,
]
coordinates {
    (100,0.871)
    (128,0.896)
    (200,0.900)
    (256,0.903)
    (300,0.907)
};
\addlegendentry{JF17K}
\addplot[
    color=red,
    thick,
    dashed
]
coordinates {
    (100,0.753)
    (128,0.759)
    (200,0.760)
    (256,0.761)
    (300,0.759)
};
\addlegendentry{WD50K}

\addplot[
    color=blue,
    thick,
    dotted
]
coordinates {
    (100,0.846)
    (128,0.854)
    (200,0.855)
    (256,0.856)
    (300,0.856)
};
\addlegendentry{Combined-HEHR}

\end{axis}
\end{tikzpicture}
 \caption{Link Prediction Performance vs. Embedding Dimension}
    \label{fig:emb_vs_mrr}
\end{figure}
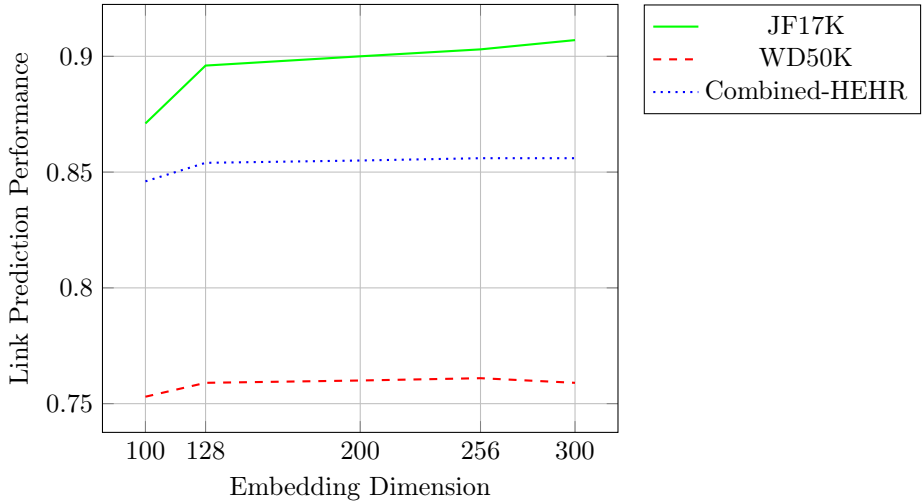

\textbf{Embedding Dimension:} The embedding dimension used for the presented results is 128 which is chosen empirically by a search across the values (100, 128, 200, 256, 300). It is observed that there is no considerable improvement in link prediction performance for higher values. A plot of link prediction performance against different embedding dimensions is given in Figure\ref{fig:emb_vs_mrr}

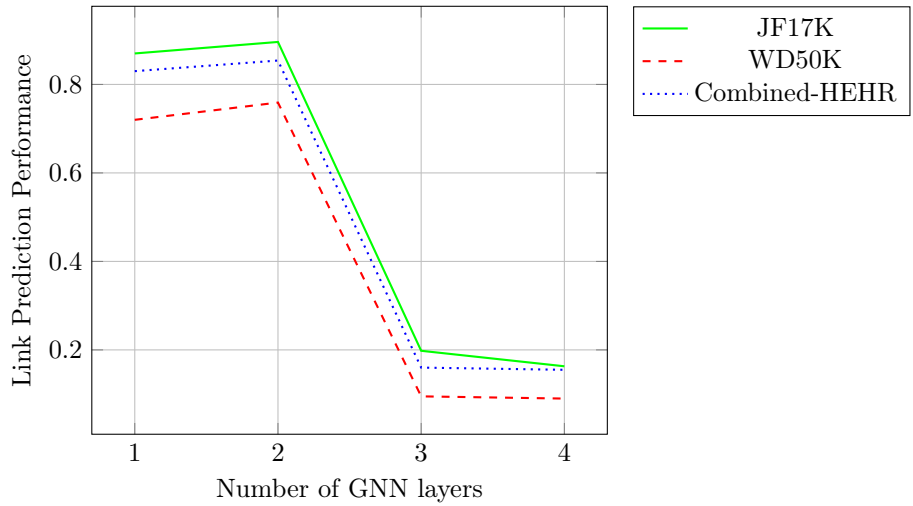
\begin{figure}[h!]
    \centering
\begin{tikzpicture}
  \begin{axis}[
    xlabel={Number of GNN layers},
    ylabel={Link Prediction Performance},
    legend style={at={(1.05,1)}, anchor=north west},
    grid=both,
    scaled ticks=false,
    xtick={1,2,3,4},
    yticklabel style={/pgf/number format/.cd,fixed,precision=3},]

\addplot[
    color=green,
    thick,
]
coordinates {
    (1,0.870)
    (2,0.896)
    (3,0.198)
    (4,0.163)
};
\addlegendentry{JF17K}
\addplot[
    color=red,
    thick,
    dashed
]
coordinates {
    (1,0.720)
    (2,0.759)
    (3,0.095)
    (4,0.090)
};
\addlegendentry{WD50K}

\addplot[
    color=blue,
    thick,
    dotted
]
coordinates {
    (1,0.830)
    (2,0.854)
    (3,0.160)
    (4,0.155)
};
\addlegendentry{Combined-HEHR}

\end{axis}
\end{tikzpicture}
 \caption{Link Prediction Performance vs. Number of GNN layers}
    \label{fig:num_gnn_vs_mrr}
\end{figure}

The number of GNN layers used is empirically fixed at 2 by comparing the performance across the values (1,2,3,4). An improvement in link prediction performance as well as a faster rate of convergence are observed when the number of GNN layers is increased from 1 to 2. However, when the number is increased to 3 or 4, the model is not able to converge unless batch normalization is used and no significant improvement is noticed in prediction performance. A plot of link prediction performance with number of GNN layers varying from 1 to 4 is given in Figure \ref{fig:num_gnn_vs_mrr}

\par \textbf{Inductive and Transductive modes:} The encoder is designed to support two modes of operation - inductive and transductive. In inductive mode, the trained parameters are independent of  any dataset specific details like number of entities or number of relations. The weights in the GNN layers and the decoder are the only learned parameters in this mode. The feature vectors for the entities and relations are initialized to a constant, which has been  empirically fixed as a vector with all values equal to 0.5. In transductive mode, the initial feature vectors for the entities and relations are also part of the learned parameters. The presented results are with the transductive mode, as it is found to give a better  performance over the inductive model. 

\subsection{Scalability to larger datasets and Analysis of Memory and Computational Complexity}
We ran experiments to empirically assess the memory and computational complexity of HEHRGNN, and to see how the model performs with larger datasets.

We constructed three large datasets primarily to demonstrate HEHRGNN's ability to scale to larger datasets with millions of edges and to understand its empirical computational complexity. These three  datasets, comprising approximately 6 million, 4 million and 2million triples, were constructed by extracting those many triples randomly from the Yago 4 dataset\cite{PellissierTanon2020YAGOBase}. The statistics of these datasets are given in Table \ref{tab:large_datasets}. 

Link prediction experiments conducted on these datasets confirm that HEHRGNN is able to scale to larger datasets with millions of facts. The link prediction result on Yago-6M is presented in Table \ref{tab:lp_results_yago}, which shows that the model is able to perform link prediction with reasonable accuracy on a dataset with 6 million facts.   

 \begin{table}[h!]
	\centering
    \vspace{-12pt}
    	 \caption{Large datasets }
	\label{tab:large_datasets}
\addtolength{\tabcolsep}{-5pt}
	\begin{tabular}{ccccccc}
		\toprule
		 Dataset & Train facts & Valid facts & Test facts & $|Entities|$ & $|Relations|$\\
		\midrule
          Yago-6M & 4463123 & 463781 & 1448986 & 5196467 & 127\\
          Yago-4M & 3708488 & 45678 & 3708488 & 4761044 & 127\\
          Yago-2M & 1817158 & 26523 & 108335 & 2502638 & 125\\
		\bottomrule
	\end{tabular}

   \end{table}

  \begin{table}[h!]
        \centering
\begin{minipage}{0.5\textwidth}
        \centering
          \caption{\shortstack{Link Prediction results on large dataset with millions of  facts}}\label{tab:lp_results_yago}
\addtolength{\tabcolsep}{-5pt}
 
        \begin{tabular}{ccccc}
                \toprule
        Model &  \multicolumn{4}{c} {Yago-6M dataset}\\
         \cmidrule(lr){2-5}
                   &  MRR & Hits@1 & Hits@3 & Hits@10\\
                \midrule
                 HEHRGNN &  0.506  & 0.315 & 0.603 & 0.956  \\
                \bottomrule
        \end{tabular}       
\end{minipage}
\end{table}
We studied the memory and compute resource usage by HEHRGNN model for different datasets. The memory and compute resource usage values recorded on Nvidia H100-94GB for an embedding dimension of 128 are presented in Table \ref{tab:mem_comp_usage}.
\REM{
\begin{figure}[h!]
    \centering
\begin{tikzpicture}
  \begin{axis}[
    xlabel={Memory usage(in GBs)},
    ylabel={Compute usage(as \%)},
    legend style={at={(1.05,1)}, anchor=north west},
    grid=both,
  ]
  \addplot[
    only marks,
    color=red,
    mark=square,
  ] coordinates {
    (1.9,30)
  };
  \addlegendentry{JF17K}

  \addplot[
    only marks,
    color=blue,
    mark=square*,
  ] coordinates {
    (3.1, 50)
  };
  \addlegendentry{WD50K}

  \addplot[
    only marks,
    color=green,
    mark=halfcircle*,
  ] coordinates {
    (4.2,55)
  };
 \addlegendentry{Combined-HEHR}
  \addplot[
    only marks,
    color=cyan,
    mark=halfdiamond*,
  ] coordinates {
    (28,85)
  };
 
  \addlegendentry{Yago-2M}
     \addplot[
    only marks,
    color=black,
    mark=otimes*,
  ] coordinates {
    (56,90)
  };
  \addlegendentry{Yago-4M}
 
   \addplot[
    only marks,
    color=magenta,
    mark=heart,
  ] coordinates {
    (89,95)
  };
  \addlegendentry{Yago-6M}
  \end{axis}
\end{tikzpicture}
 \caption{Memory and Compute usage by HEHRGNN for different datasets(with embedding dimension of 128)}
    \label{fig:mem_compute_datasets}
\end{figure}
}

 \begin{table}[h!]
	\centering
    	 \caption{Memory and Compute usage by HEHRGNN for different datasets(with embedding dimension of 128)}
	\label{tab:mem_comp_usage}
\addtolength{\tabcolsep}{-5pt}
\resizebox{\textwidth}{!}{%
	\begin{tabular}{ccccccc}
		\toprule
		   &  \multicolumn{6}{c} {Dataset}\\
         \cmidrule(lr){2-7}
         &JF17K&WD50K&Combined-HEHR&Yago-2M&Yago-4M&Yago-6M\\
         \midrule
          Memory(GB) & 1.9&3.1&4.2&28&56&89\\
          Compute(\% utilization) & 30&50&55&85&90&95\\
		\bottomrule
	\end{tabular}
}
   \end{table}

 We also compared the memory and compute usage of HEHRGNN with that of the publicly available implementations of two baseline models - STaRE \footnote{\url{https://github.com/migalkin/StarE}} and HyPE\footnote{\url{https://github.com/ServiceNow/HypE}}. 
The values of resource usage by the different models are given in Table  \ref{tab:compare_mem_comp_usage}. The STARE model could not be run with Yago-6M on Nvidia H100-94GB for want of memory.
\REM{
\begin{figure}[h!]
    \centering
\begin{tikzpicture}
  \begin{axis}[
    xlabel={Memory usage(in GBs)},
    ylabel={Compute usage(\%)},
    legend style={at={(1.05,1)}, anchor=north west},
    grid=both,
  ]

  \addplot[
    only marks,
    color=blue,
    mark=triangle*,
  ] coordinates {
    (3.1, 50)
  };
  \addlegendentry{HEHRGNN (WD50K) }

 \addplot[
    only marks,
    color=red,
    mark=triangle*,
  ] coordinates {
    (5.8, 70)
  };
  \addlegendentry{STARE (WD50K) }

   \addplot[
    only marks,
    color=Blue,
    mark=square*,
  ] coordinates {
    (1.9,30)
  };
  \addlegendentry{HEHRGNN (JF17K) }

  \addplot[
    only marks,
    color=cyan,
    mark=square*,
  ] coordinates {
    (1.05,25)
  };
  \addlegendentry{HyPE (JF17K) }

 \addplot[
    only marks,
    color=cyan,
    mark=diamond*,
  ] coordinates {
    (9,90)
  };
  \addlegendentry{HyPE (Yago-4M) }

\addplot[
    only marks,
    color=blue,
    mark=diamond*,
  ] coordinates {
    (56,90)
  };
  \addlegendentry{HEHRGNN (Yago-4M) }

  \addplot[
    only marks,
    color=red,
    mark=diamond*,
  ] coordinates {
    (80,100)
  };
  \addlegendentry{STARE (Yago-4M) }

 \addplot[
    only marks,
    color=cyan,
    mark=*,
  ] coordinates {
    (14,100)
  };
  \addlegendentry{HyPE (Yago-6M) }

\addplot[
    only marks,
    color=blue,
    mark=*,
  ] coordinates {
    (89,95)
  };
  \addlegendentry{HEHRGNN (Yago-6M) }  
  \end{axis}
\end{tikzpicture}
\caption{Comparison of Memory and Compute usage by HEHRGNN and the two baselines}
    \label{fig:mem_compute_baselines}
\end{figure}
}
 \begin{table}[h!]
	\centering
    	 \caption{Comparison of Memory and Compute usage by HEHRGNN and the two baselines}
	\label{tab:compare_mem_comp_usage}
\addtolength{\tabcolsep}{-2pt}
\resizebox{\textwidth}{!}{%
	\begin{tabular}{ccccccccccc}
		\toprule
		   Dataset&  \multicolumn{2}{c} {WD50K} & \multicolumn{2}{c}{JF17K} & \multicolumn{3}{c}{Yago-4M} & \multicolumn{3}{c}{Yago-6M}\\
         \cmidrule(lr){2-3} \cmidrule(lr){4-5} \cmidrule(lr){6-8} \cmidrule(lr){9-11}         Model&HEHRGNN&STARE&HypE&HEHRGNN&HypE&HEHRGNN&STARE&HypE&HEHRGNN&STARE\\
         \midrule
          Memory(GB) & 3.1 & 5.8 & 1.1 & 1.9 &9&56&80&14&89&-\\
          Compute(\% utilization) & 50&70&25&30&90&90&100&100&95& -\\
		\bottomrule
	\end{tabular}
}
   \end{table}
It can be observed from the figure that the computational and memory complexity of HEHRGNN is better compared to STARE on WD50K. We believe it is mainly because STARE is using Transformer based decoder whereas HEHRGNN uses DistMult or HyPE decoder functions.

The main compute operations involved in embedding generation in HEHRGNN are i) the hyperedge centric propagation for the Gather step, ii) the scatter-reduce operation for updating the relation embeddings from edge instances in the Apply step, and iii) the node-centric propagations in the Scatter step. As these operations are parallelized by the PyTorch backend, HEHRGNN is able to leverage the compute resources efficiently for embedding generation. 

\subsection{Ablation studies}
The main distinguishing aspects of the model are the knowledge graph handling structure that supports hyper-relational edges (edges with qualifier pairs) as well as hyperedges (edges connecting more than two nodes), and the GNN-based encoder with a novel propagation mechanism that supports both of these edge types. We conducted ablation experiments to understand the effectiveness of these model design factors in terms of embedding quality. 
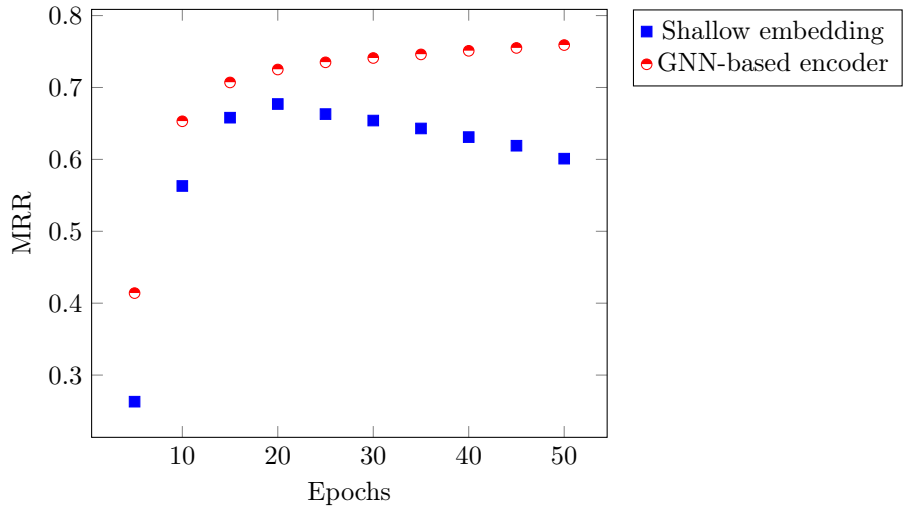
\begin{figure}[h!]
    \centering
\begin{tikzpicture}
  \begin{axis}[
    xlabel={Epochs},
    ylabel={MRR},
    legend style={at={(1.05,1)}, anchor=north west},
    legend entries=
    grid=both,
  ]
  \addplot[
    only marks,
    color=blue,
    mark=square*,
  ] coordinates {
    (5,0.263)
    (10,0.563)
    (15,0.658)
    (20,0.677)
    (25,0.663)
    (30,0.654)
    (35,0.643)
    (40,0.631)
    (45,0.619)
    (50,0.601)
  };
  \addlegendentry{Shallow embedding}

 \addplot[
    only marks,
    color=red,
    mark=halfcircle*,
      ] coordinates {
    (5,0.414)
    (10,0.653)
    (15,0.707)
    (20,0.725)
    (25,0.735)
    (30,0.741)
    (35,0.746)
    (40,0.751)
    (45,0.755)
    (50,0.759)   
  };
  \addlegendentry{GNN-based encoder}

  \end{axis}
\end{tikzpicture}
\caption{Training curves of GNN-based encoder and Shallow embedding for WD50K dataset}
    \label{fig:gnn_vs_shallow}
\end{figure}
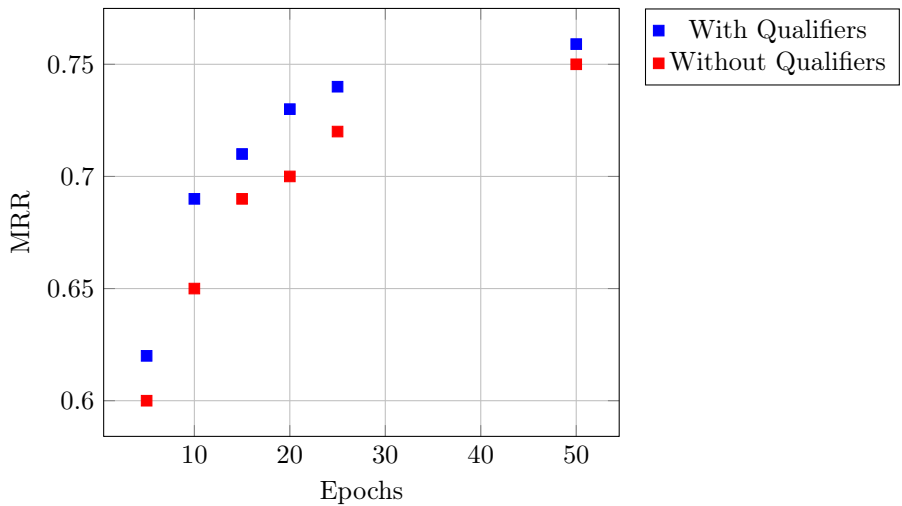
\begin{figure}[H]
    \centering
\begin{tikzpicture}
  \begin{axis}[
    xlabel={Epochs},
    ylabel={MRR},
    legend style={at={(1.05,1)}, anchor=north west},
    grid=both,
  ]
  \addplot[
    only marks,
    color=blue,
    mark=square*,
  ] coordinates {
    (5,0.620)
    (10,0.690)
    (15,0.710)
    (20,0.730)
    (25,0.740)
    (50,0.759)
  };
  \addlegendentry{With Qualifiers}

 \addplot[
    only marks,
    color=red,
    mark=square*,
  ] coordinates {
    (5,0.600)
    (10,0.650)
    (15,0.690)
    (20,0.700)
    (25,0.720)
    (50,0.750)    
  };
  \addlegendentry{Without Qualifiers}
  \end{axis}
\end{tikzpicture}
\caption{Comparison of MRR growth when WD50K is processed with and without qualifiers}
    \label{fig:mrr_with_without_qual}
\end{figure}
\textbf{GNN-based encoder vs. Shallow embedding:} To study the effectiveness of the proposed GNN-based encoder,  we conducted experiments by disabling the encoder part and running the model in a shallow embedding mode. All the remaining parts, including the decoder, were retained as in the original model. In this case, the entity and relation embedding matrices were directly optimized based on the decoder scores. The MRR values against the epochs for both the cases are plotted in Figure \ref{fig:gnn_vs_shallow}. It can be seen that the proposed GNN-based encoder is achieving a significantly better accuracy and also has a stable training curve. This experiment results indicate  that the performance improvement of HEHRGNN over HypE is mainly due to the use of the GNN-based encoder in the embedding model. HypE is a decoder only model equivalent to the shallow embedding mode of HEHRGNN.

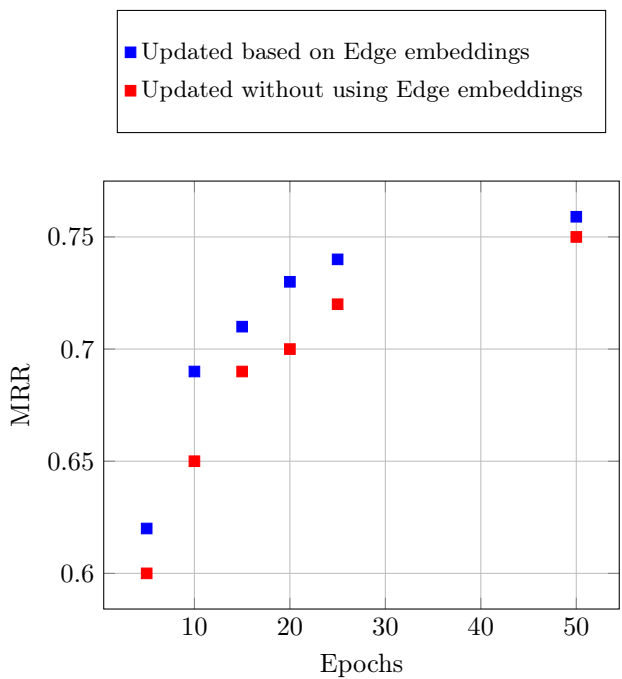
\begin{figure}[h!]
    \centering
\begin{tikzpicture}
  \begin{axis}[
    xlabel={Epochs},
    ylabel={MRR},
    legend style={text width=6cm, align=left,font=\small, row sep=0.1cm, inner ysep=10pt,at={(0.5,1.4)}, anchor=north },
    grid=both,
  ]

  \addplot[
    only marks,
    color=blue,
    mark=square*,
  ] coordinates {
    (5,0.620)
    (10,0.690)
    (15,0.710)
    (20,0.730)
    (25,0.740)
    (50,0.759)
  };
  \addlegendentry{Updated based on Edge embeddings}

 \addplot[
    only marks,
    color=red,
    mark=square*,
  ] coordinates {
    (5,0.600)
    (10,0.650)
    (15,0.690)
    (20,0.700)
    (25,0.720)
    (50,0.750)    
  };
  \addlegendentry{Updated without using Edge embeddings } 
  \end{axis}
\end{tikzpicture}
\caption{Comparison of MRR growth for the two different methods of updating relation embeddings}
    \label{fig:rel_with_without_edge}
\end{figure}

 \textbf{Model performance with and without the qualifier processing part:} We carried out a set of experiments to study the impact of the qualifier processing part of the model. In the first experiment, we compared the MRR convergence of the model with respect to three subsets of WD50K with different ratios of facts with qualifiers. These subsets, namely WD50K\_100 (all the facts having qualifiers), WD50K\_66(66 percent of facts having qualifier,s) and WD50K\_33(33 percent of facts having qualifiers) ,were used in \cite{galkin2020message}. The convergence of the model (an MRR of 0.7 was set as a reasonable threshold for easy comparison of convergence)  is found to be faster in datasets with higher qualifier ratio. This is illustrated in the Table \ref{tab:mrr_conv} with the number of sample propagations required for attaining the threshold MRR value for the different subset datasets. The second experiment was to compare the MRR growth with respect to the number of epochs when the WD50K dataset is processed with qualifiers and without qualifiers. The MRR growth is found to be slower when processed without qualifiers, as shown in Figure \ref{fig:mrr_with_without_qual}, confirming the importance of qualifier.  
These two experiments ascertain that the qualifier processing part in our model contributes significantly to the quality of embedding generation.  

\REM{
 \vspace{-3pt}
 \begin{table}[h!]
	\centering
    \vspace{-12pt}
    	 \caption{Convergence on datasets with different ratios of qualifiers}
	\label{tab:mrr_conv}
\addtolength{\tabcolsep}{-5pt}
	\begin{tabular}{cccc}
		\toprule
		 Dataset & No. of facts & No. of Epochs & No. of forward propagations for MRR threshold\\
		\midrule
		WD50K\_33 & 73406 & 15 & 1101090\\
        WD50K\_66 & 35968 & 22 & 791296 \\
   WD50K\_100 & 22738 & 14 & 318332 \\
		\bottomrule
	\end{tabular}
   \end{table}
}
\vspace{-3pt}
 \begin{table}[h!]
	\centering
    \vspace{-12pt}
    	 \caption{Convergence on datasets with different ratios of qualifiers}
	\label{tab:mrr_conv}
\addtolength{\tabcolsep}{-5pt}
\resizebox{\textwidth}{!}{%
	\begin{tabular}{|p{3cm}|p{4.8cm}|p{6cm}|}
		\toprule
		 Dataset &  Sample propagations for attaining MRR threshold & Calculation of sample propagations (No of facts in dataset x Epochs for threshold)\\
		\midrule
		WD50K\_33 &  1101090 & 73406 x 15\\
        WD50K\_66 &  791296 & 35968 x 22 \\
   WD50K\_100 &  318332 & 22738 x 14\\
		\bottomrule
	\end{tabular}
    }
   \end{table}
 
The next study was to see the effect of the propagation mechanism in the model. The model uses scattering of edge instance embeddings to update the relation
embeddings. This was disabled and in turn a simple learned weight matrix was used to directly transform and update the relation
embeddings. The MRR growth was found to be slower, as shown in the plot in Figure \ref{fig:rel_with_without_edge} confirming the significance of the design.

\vspace{-0.2in}
\section{Conclusion and Future Work}
We discussed relational hyperedges and hyper-relational edges which are essential in \kgname frameworks for capturing the commonly occurring n-ary relational facts in real-world.  We proposed HEHRGNN, a GNN-based unified embedding model for knowledge graphs with both hyperedges and hyper-relational edges. The experimental evaluations  show the effectiveness of the model for link prediction on individual hyperedge and hyper-relational datasets, as well as on combined datasets. A unified fact representation format for datasets is also proposed to enable maintaining and processing common datasets for link prediction. The scalability of the model is also demonstrated with a large dataset of 6 Million facts.  We see that this work can be extended in multiple directions including: investigating GNN architectures and layer depth for improving the inductive performance across different datasets,  building good representative datasets in the unified format covering binary and n-ary relations.\\~\\  
\textbf{Broader impact} We hope this work helps kindle interest towards unification of two currently separate lines of research - knowledge hypergraphs and hyper relational knowledge graphs- into one. Unified models will improve the efficacy of knowledge graphs by increasing their expressiveness to capture the real-world facts  in their entirety.
\bibliographystyle{elsarticle-num} 
\bibliography{reference.bib}



\end{document}